\documentclass{article}
\usepackage{arxiv}

\usepackage[utf8]{inputenc}
\usepackage[T1]{fontenc}
\usepackage{microtype}
\usepackage{graphicx}
\usepackage{url}
\usepackage{hyperref}
\usepackage{natbib}

\usepackage{booktabs}
\usepackage{array}
\usepackage{tabularx}
\usepackage{siunitx}

\usepackage{float}      
\usepackage{adjustbox}  
\usepackage{tikz}
\usetikzlibrary{arrows.meta,positioning,calc}

\usepackage{amsfonts}
\usepackage{nicefrac}
\usepackage{lipsum}   

\title{PolyTruth: Multilingual Disinformation Detection using Transformer-Based Language Models}

\renewcommand{\shorttitle}{Multilingual Disinformation Detection with Transformers}

\author{%
\href{https://orcid.org/0009-0008-6588-6945}{\includegraphics[scale=0.06]{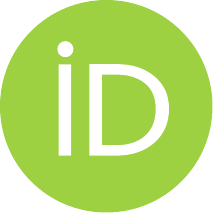}\hspace{1mm}Zaur Gouliev} \\
School of Information \& Communication Studies \\
University College Dublin, Ireland \\
\texttt{zaur.gouliev@ucdconnect.ie} \\
\And
\href{https://orcid.org/0009-0009-7921-4741}{\includegraphics[scale=0.06]{orcid.pdf}\hspace{1mm}Chengqian Wang} \\
School of Information \& Communication Studies \\
University College Dublin, Ireland \\
\texttt{chengqian.wang@ucdconnect.ie} \\
\And
\href{https://orcid.org/0009-0002-0683-2978}{\includegraphics[scale=0.06]{orcid.pdf}\hspace{1mm}Jennifer Waters} \\
School of Information \& Communication Studies \\
University College Dublin, Ireland \\
\texttt{jennifer.waters@ucdconnect.ie} \\
}

\hypersetup{
pdftitle={PolyTruth: Multilingual Disinformation Detection using Transformer-Based Language Models},
pdfsubject={cs.CL, cs.ML},
pdfauthor={Zaur Gouliev, Chengqian Wang, Jennifer Waters},
pdfkeywords={Disinformation Detection, Multilingual NLP, Transformer Models, PolyTruth Disinformation Corpus, Cross-Lingual Evaluation},
}

\begin{document}
\maketitle

\begin{abstract}
Disinformation spreads rapidly across linguistic boundaries, yet most AI models are still benchmarked only on English. We address this gap with a systematic comparison of five multilingual transformer models: mBERT, XLM, XLM-RoBERTa, RemBERT, and mT5 on a common fake-vs-true machine learning classification task. While transformer-based language models have demonstrated notable success in detecting disinformation in English, their effectiveness in multilingual contexts still remains up for debate. To facilitate evaluation, we introduce PolyTruth Disinfo Corpus, a novel corpus of 60,486 statement pairs (false claim vs. factual correction) spanning over twenty five languages that collectively cover five language families and a broad topical range from politics, health, climate, finance, and conspiracy, half of which are fact-checked disinformation claims verified by an augmented MindBugs Discovery dataset. Our experiments revealed performance variations. Models such as RemBERT achieved better overall accuracy, particularly excelling in low-resource languages, whereas models like mBERT and XLM exhibit considerable limitations when training data is scarce. We provide a discussion of these performance patterns and implications for real-world deployment. The dataset is publicly available on our GitHub repository to encourage further experimentation and advancement. Our findings illuminate both the potential and the current limitations of AI systems for multilingual disinformation detection.
\end{abstract}


\section{Introduction}
Online disinformation has become pervasive, spreading rapidly across social media and online platforms. Studies have shown that false news can spread faster and farther than the truth in online networks~\cite{vosoughi2018spread}, exacerbating the challenge for automated systems. Disinformation detection has thus attracted significant research attention in recent years~\cite{shu2017fake}. However, much of the earlier work focused on English-language content, taking advantage of large labeled datasets and advances in deep learning for text classification \cite{supriyono2024nlp}. The multilingual dimension of the disinformation problem remains relatively under-explored due to the scarcity of annotated data in languages beyond English~\cite{hardalov2021survey,chalehchaleh2024multilingual}. One of the broad motivations of this study was the Facebook (Meta) scandal in Myanmar and system-level inability of its algorithms to filter out harmful content because it didn't have enough training data on the language \cite{amnesty2022myanmar,zaleznik2021facebook}. There are many reasons for this, but one reasons is that building robust multilingual disinformation detectors is inherently difficult, and false claims often propagate in multiple languages, including low-resource languages where detection tools are limited~\cite{hammouchi2022evidence}. Multilingual transformer-based language models offer a potential solution for this task. Models such as mBERT~\cite{devlin2019bert} and XLM-RoBERTa~\cite{conneau2020xlmr} learn joint representations for dozens of languages and have demonstrated impressive cross-lingual transfer capabilities on tasks like question answering. Recent studies have started to explore multilingual detection using such models~\cite{chalehchaleh2024multilingual}, and have found that multilingual training strategies can substantially improve performance. However, in the literature, comparisons of different multilingual transformers on a unified disinformation detection task are lacking. In this paper, we address this gap by comparing five state-of-the-art multilingual transformer models on the task of disinformation statement classification. We make use of the MindBugs Discovery dataset~\cite{cheres2024mindbugs}, which contains 30,243 debunked disinformation statements from Europe (2009–2024) in various languages. Each statement is annotated with its veracity (false/disinformation) and language. We augmented and created a complimentary dataset with true (non-disinformation) statements to enable binary classification.\footnote{For each false claim, we paired it with a corresponding true statement or a factual correction from fact-checking sources to serve as a negative example. This yielded a balanced dataset for training.} We fine-tune and evaluate five transformer-based language models: BERT-base-multilingual-cased (mBERT)~\cite{devlin2019bert}, XLM (Cross-lingual Language Model)~\cite{lample2019xlm}, XLM-RoBERTa (base)~\cite{conneau2020xlmr}, RemBERT~\cite{chung2021rembert}, and mT5~\cite{xue2021mt5} under the same experimental conditions. Our evaluation focuses on two key aspects, the overall model performance on the multilingual data, and performance differences between high-resource and low-resource language subsets.

Our contributions are as follows: \textbf{(1)} We expand prior work by conducting a thorough comparative evaluation of five multilingual transformer models on a common disinformation detection task, providing insights into their strengths and weaknesses across languages. \textbf{(2)} We describe a unified training and evaluation framework for multilingual fake news detection, including dataset preparation and preprocessing steps that can serve as a reference for future research. \textbf{(3)} We analyse model performance across languages, highlighting how high-resource languages benefit from abundant data and how low-resource languages pose challenges where certain models (notably XLM-R and RemBERT) still perform robustly. To our knowledge, this is a comprehensive evaluation of multilingual transformers for disinformation detection. We hope our findings will inform the development of more effective multilingual and cross-lingual fake news detection systems.

The remainder of the paper is organised as follows. Section~\ref{sec:related} reviews related work in multilingual disinformation detection. Section~\ref{sec:methods} describes the dataset and our methodology, including preprocessing and model fine-tuning setup. Section~\ref{sec:experiments} presents the experimental results, comparing model performances overall and by language groups, with analysis. Finally, Section~\ref{sec:conclusion} concludes the paper and outlines future work.

\section{Related Work}\label{sec:related}
Early research on disinformation and fake news detection largely centered on English content, using machine learning and deep learning to classify news articles or social media posts as fake or real~\cite{shu2017fake}. Traditional approaches relied on textual features like n-grams, readability, and linguistic cues or even incorporated user and network metadata. With the rise of deep neural models, many works applied CNNs, RNNs, or attention-based architectures to detect disinformation, showing improved accuracy on benchmark datasets. However, these efforts often did not generalise to other languages due to language-specific patterns and resource disparities. Addressing multilingual detection presents a unique challenge because labeled data in most languages are scarce. One line of work has focused on cross-lingual transfer, where a model trained on a high-resource language is applied to detect fake news in low-resource languages. Studies in the past have translated non-English data into English to use existing detectors, or conversely translated English training data into the target language to fine-tune models~\cite{chalehchaleh2024multilingual}. While these translation-based strategies leverage English resources, they risked meaning loss and translation errors, especially for low-resource languages. Another approach was to use multilingual transformers pre-trained on multiple languages. These models could directly encode text in different languages into a shared semantic space, enabling zero-shot or few-shot transfer. The multilingual BERT model (mBERT)~\cite{devlin2019bert} was one of the first to show cross-lingual capabilities, despite being trained without explicit alignment objectives, mBERT can often generalise to languages unseen in fine-tuning. Lample and Conneau’s XLM model~\cite{lample2019xlm} improved cross-lingual performance by introducing a translation language modeling objective to better align representations across languages. The XLM-RoBERTa (XLM-R) model~\cite{conneau2020xlmr} further advanced the state of the art by training on a massive multilingual corpus known as Common Crawl data containing 100 languages, and with a larger model size and improved pre-training, this led to significant gains on tasks like cross-lingual natural language inference and question answering. 

These multilingual transformers have been applied to disinformation tasks in recent years and challenges have been put out to the NLP communities to address this. The CheckThat! 2022 Lab included a cross-lingual fake news detection task (English to German) where participants successfully fine-tuned XLM-R for transfer learning~\cite{schutzecheckthat2022}. This built on Patwa et al.~\cite{patwa2021constraint}, in the CONSTRAINT 2021 shared task on fake news detection in English and Hindi, demonstrating that transformers can handle multilingual and even code-switched content given appropriate fine-tuning. Further research conducted on various models and training scenarios \cite{chalehchaleh2024multilingual} experimented with monolingual vs. multilingual training, noting that training on combined multilingual data can improve low-resource language performance, albeit with some trade-offs for high-resource languages due to class imbalance. In follow-up work, Chalehchaleh et al.~\cite{chalehchaleh2025llm} also looked at large language model-based data augmentation to enhance multilingual fake news detection by generating synthetic training examples for low-resource languages using GPT-style models, reporting improved performance and highlighting data augmentation as a viable solution to the low-resource problem. This is in line with our work. Lesser explored advancements have also looked the integration of external evidence into multilingual disinformation detection systems. Hammouchi and Ghogho~\cite{hammouchi2022evidence} proposed an evidence-aware multilingual fake news detection framework, which combines multilingual transformer models with evidence retrieval and source credibility mechanisms. This approach, evaluated on COVID-19 disinformation across multiple languages, showed significant accuracy improvements (e.g., achieving an F1-score of 0.85 on the XFACT dataset). Another method shown by Zhou et al.~\cite{zhou2023crosslingual} explored cross-lingual knowledge transfer techniques, demonstrating that leveraging related languages during fine-tuning can substantially enhance performance on low-resource language disinformation tasks. A final promising direction is the use of integrated multi-modal signals \cite{gupta2024multimodal}, such as combining images and text, for multilingual fake news detection, this demonstrated that multimodal models can significantly outperform text-only baselines, particularly useful as they could be employed by very large online platforms due to how visually rich they are.

Despite these developments, the literature still lacks extensive head-to-head comparisons of multiple multilingual transformer models on disinformation detection tasks. Most existing studies evaluate a single model or only a few models in isolation. We aim to address this gap by systematically evaluating five multilingual transformers under identical conditions, providing insights into their relative strengths and limitations. Our work differs from evidence-based approaches in that we focus on content-based detection, where the models make predictions solely from the statement text. This setup allows us to directly assess the language understanding and generalisation capability of multilingual transformers on the task. This can also help identify which architectures are most suitable for multilingual disinformation detection on similar datasets.

\section{Data and Methodology}\label{sec:methods}

\subsection{Dataset and Preparation}
\begin{figure}[t!]
  \centering
  \includegraphics[width=\textwidth]{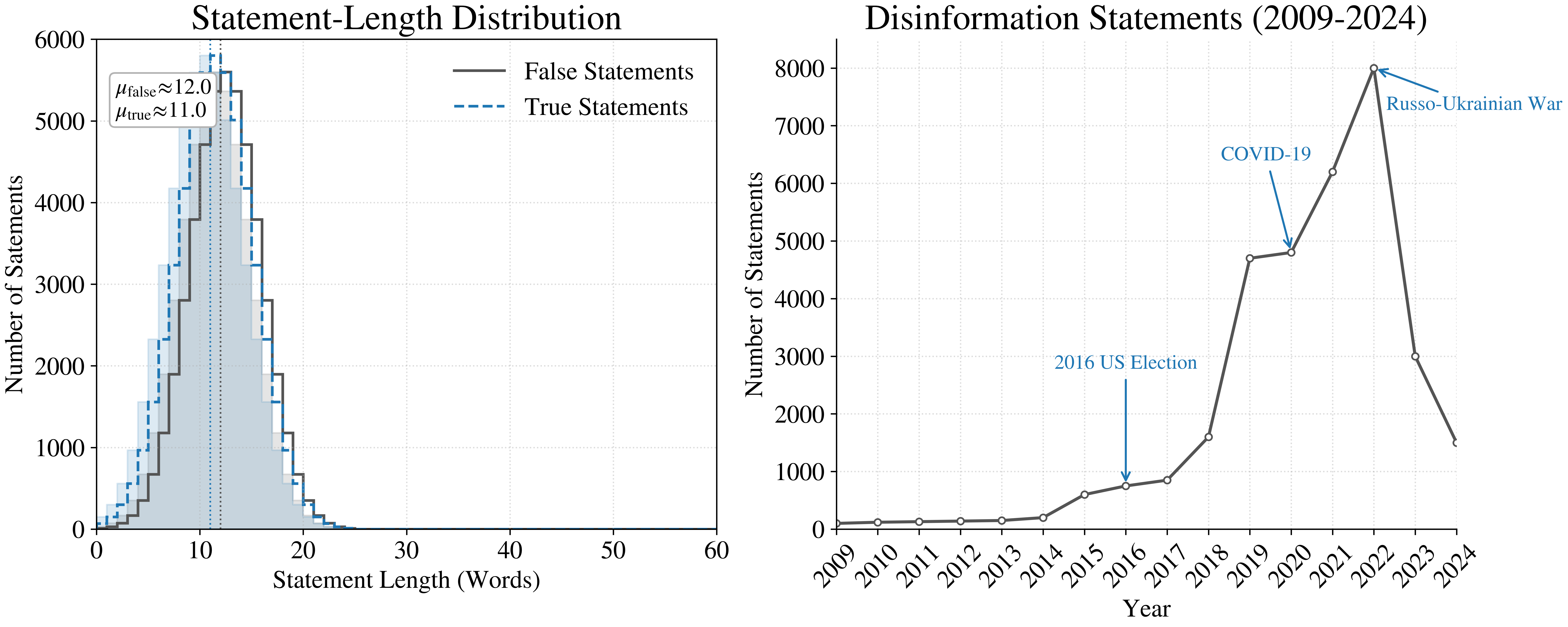}
  \caption{Left—\textbf{Statement length}: both false and true claims peak at 10–15 words; false statements exhibit a slightly longer tail. 
           Right—\textbf{Yearly volume}: counts stay low until 2015, rise with the \emph{2016 US election}, spike during the \emph{COVID-19} infodemic (2020) and surge again with the \emph{2022 Russo-Ukrainian war}.}
  \label{fig:length-and-volume}
\end{figure}

We base our experiments on a multilingual disinformation dataset provided as part of the MindBugs Discovery project~\cite{cheres2024mindbugs}. The dataset consists of 30,243 statements that have been identified as disinformation (fake or misleading claims) by verified fact-checking organizations in Europe between 2009 and 2024. Each data instance includes the text of the statement, the language of the statement, the date it was debunked, and other metadata such as the fact-check source. The statements cover a wide range of topics from political falsehoods, health disinformation to conspiracy theories, which is reflective of real-world disinformation content circulating in different countries and languages. Figure~\ref{fig:length-and-volume} shows some insights into this dataset.

For our task, we treat fake (disinformation) detection as a binary classification problem: given a statement, predict whether it is \textit{disinformation} or \textit{legitimate}. Since the collected statements are all debunked false claims, we constructed a complementary set of true or factual statements to act as the negative class called PolyTruth. Specifically, for each false statement in the dataset, we generated a corresponding true statement using OpenAI's API. Our annotation guidelines, prompt design, filtering, and quality control can also be found on this repository. These state the corrected fact, or a factual claim from a reliable source on the same topic, creating a counter-claim to the disinformation statement. In cases where a direct corrected statement was not available, we selected a truthful statement from news articles in the same language and topic. This process resulted in a balanced dataset with an equal number of true vs. false statements. After data augmentation and pairing, our full dataset for training and evaluation comprised approximately 60,486 statements (half disinformation, half true information) across all languages, consisting of 25+ distinct languages in the dataset. Table~\ref{tab:polytruth-multilang} shows an example of some statements. The \href{https://github.com/gouliev/polytruth}{\textsc{PolyTruth} corpus} is publicly available on GitHub.

\begin{table}[t]
  \centering
  \caption{Sample \textit{PolyTruth} statements in six languages}
  \label{tab:polytruth-multilang}
  \small
  \setlength{\tabcolsep}{4pt}
  \renewcommand{\arraystretch}{1.12}
  \begin{tabularx}{\linewidth}{@{}p{2.15cm} >{\raggedright\arraybackslash}X >{\raggedright\arraybackslash}X@{}}
    \toprule
    \textbf{Language} & \textbf{Disinformation statement} & \textbf{Corrective (true) statement} \\
    \midrule
    English  &
      BBVA texts customers, asking them to click a link to unblock “suspicious activity”. &
      BBVA never requests unblocking via links; such SMS are confirmed phishing attempts. \\[2pt]

    Spanish  &
      “España prohibirá el uso de hornos de gas en hogares a partir de 2025”. &
      El Gobierno sólo estudia ayudas para electrodomésticos eficientes; no existe tal prohibición. \\[2pt]

    German   &
      Die WHO plane ab 2024 einen weltweiten Impfpass, sonst seien Reisen verboten. &
      Es gibt nur eine Empfehlung für freiwillige digitale Zertifikate; ein Reiseverbot ist nicht beschlossen. \\[2pt]

    Romanian &
      Guvernul va introduce o taxă de 5\,\% pentru fiecare tranzacție bancară online în 2025. &
      Ministerul Finanțelor dezminte: nu există niciun proiect de lege pentru o astfel de taxă. \\[2pt]

    Estonian &
      Tallinna kraanivesi sisaldab ohtlikku pliid; enne joomist tuleks vesi keeta. &
      Tallinna Vesi ja Terviseamet kinnitavad, et pliisisaldus on normi piires ja vesi on joogikõlbulik. \\[2pt]

    Latvian  &
      Latvija atcels eiro un 2025.\ gadā ieviesīs jaunu digitālo valūtu “LVD”. &
      Finanšu ministrija skaidro, ka nekādu plānu par eiro atcelšanu un “LVD” ieviešanu nav. \\
    \bottomrule
  \end{tabularx}
\end{table}

\begin{figure}[!htb]
\centering
\begin{adjustbox}{max width=\linewidth}
\begin{tikzpicture}[
  font=\scriptsize,
  node distance = 5mm,
  process/.style = {rectangle, rounded corners=2pt, draw=black,
                    align=center, minimum width=3.0cm, minimum height=6mm},
  datastore/.style = {process, fill=gray!15},
  model/.style   = {process, fill=cyan!12, minimum width=2.4cm},
  misc/.style    = {process, fill=orange!15},
  arrow/.style   = {thick, -{Straight Barb[length=1.6mm,width=1mm]}},
]

\node[datastore] (raw) {Raw debunked\\statements (30\,243)};
\node[process, below=of raw] (clean) {Cleaning \& normalisation\\\tiny lower-case, strip URLs/mentions};
\node[process, below=of clean] (pair) {Pair with true statements};
\node[process, below=of pair]  (tok)  {Sub-word tokenisation};
\node[process, below=of tok]   (split){Stratified 80/10/10 split};
\node[misc, dashed, below=of split] (tune){Hyper-parameter search};

\node[model, below=10mm of tune] (xlmr){XLM-R};
\node[model, left =14mm of xlmr] (mbert){mBERT};
\node[model, right=14mm of xlmr] (xlm)  {XLM};
\node[model, below=9mm of mbert] (rem)  {RemBERT};
\node[model, below=9mm of xlm]   (mt5)  {mT5};

\node[process, below=13mm of xlmr] (eval)
      {Evaluation\\Accuracy \& F1};
\node[misc, below=of eval] (report) {Metrics report};

\foreach \a/\b in {raw/clean, clean/pair, pair/tok, tok/split}
  \draw[arrow] (\a) -- (\b);
\draw[dashed,arrow] (split) -- (tune);
\draw[arrow] (eval) -- (report);

\draw[arrow] (tune) |- (mbert.north);
\draw[arrow] (tune) |- (xlm.north);    
\draw[arrow] (tune) |- (rem.north);
\draw[arrow] (tune) |- (mt5.north);

\coordinate (midfan) at ($(tune.south)+(4mm,0)$);
\draw[dashed,arrow] (tune) -- (midfan) -- ++(0,-5mm) -| (xlmr.north);

\draw[arrow] (mbert.south) -- ++(0,-4mm) -| ($(eval.north)+(-14mm,0)$);
\draw[arrow] (xlmr.south)  -- (eval.north);
\draw[arrow] (xlm.south)   -- ++(0,-4mm) -| ($(eval.north)+(14mm,0)$);
\end{tikzpicture}
\end{adjustbox}
\caption{End‐to‐end pipeline for multilingual disinformation detection.}
\label{fig:pipeline}
\end{figure}
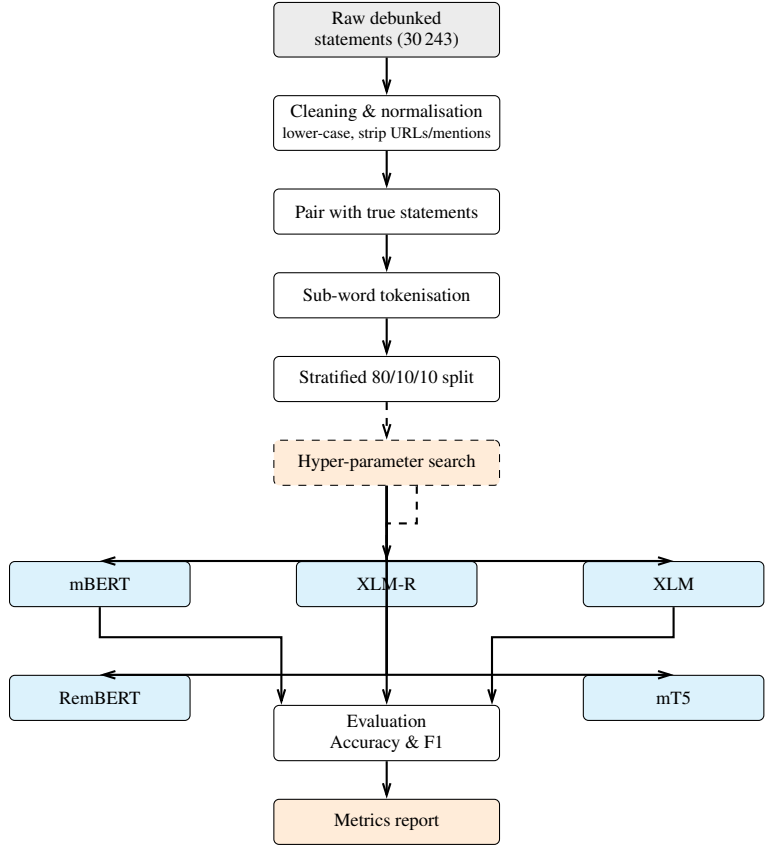

\begin{figure}[t]
  \centering
  \includegraphics[width=\textwidth]{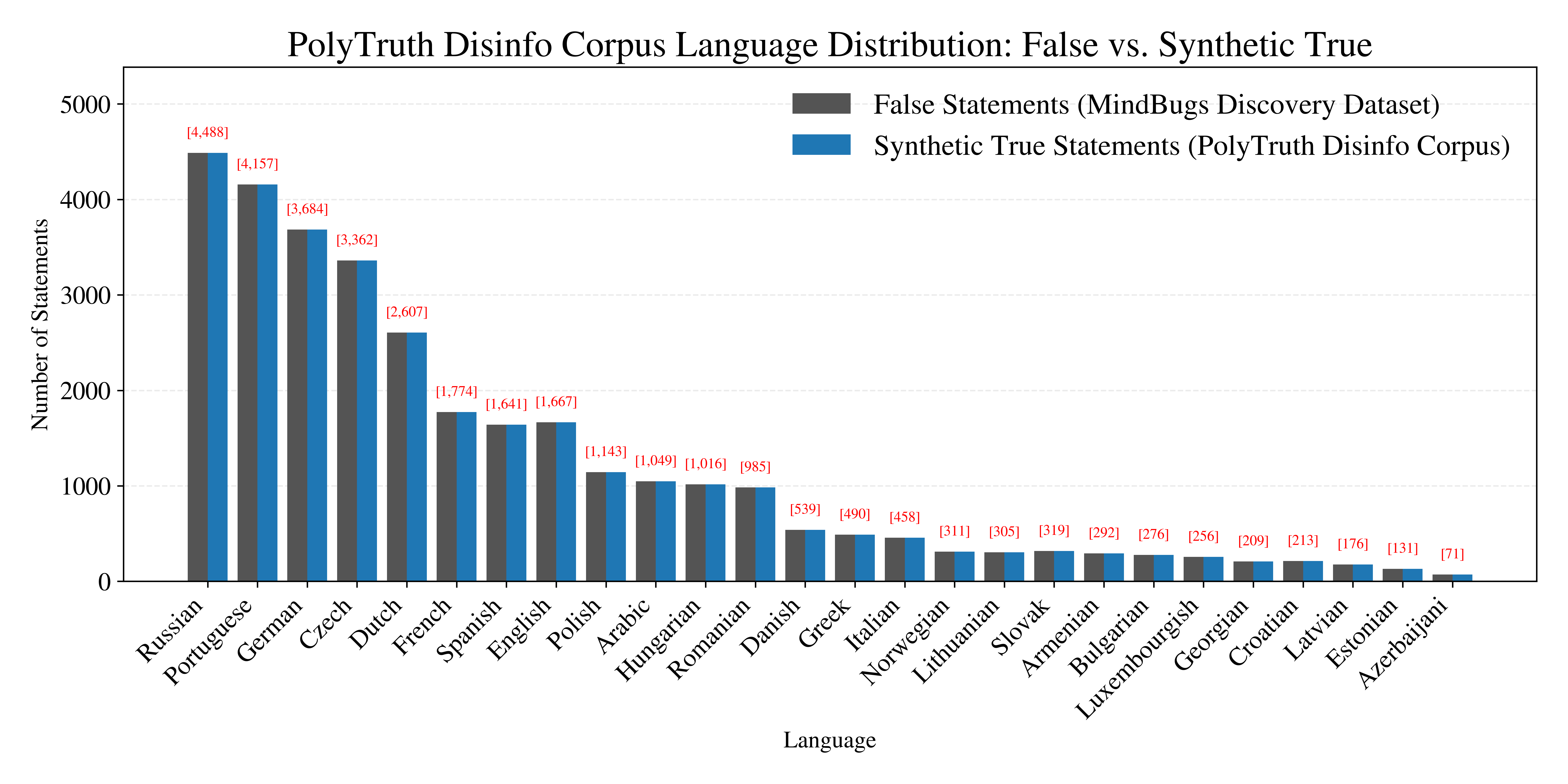}
  \caption{PolyTruth disinformation corpus: number of \emph{false} (MindBugs) and
           \emph{synthetic true} statements per language.
           The five largest languages are Russian, Portuguese, German and Czech each exceed 3,000 statements, while languages such as Latvian, Slovak, Estonian and Azerbaijani have less than a few hundred. This highlights the pronounced imbalance across languages. We regard the ten languages with at least 1,000 statements (Russian through Hungarian) as \emph{high-resource}. All remaining languages, from Romanian down to Azerbaijani, form the \emph{low-resource} group. This division underpins our subsequent analysis of how model performance varies with data availability.}
  \label{fig:lang-distribution}
\end{figure}

The dataset includes a diverse representation of languages, with Russian, Portuguese, German, and Czech being the most represented languages, each containing over 3,000 statements. Notably, Russian has the greatest representation, with approximately 4,488 false statements and a corresponding number of true pairs. Other languages such as Dutch, French, Spanish, and English each contribute between 1,500 to 3,000 statements. Several medium-resource languages like Polish, Arabic, Hungarian, and Romanian have around 1,000 statements each. Conversely, the dataset also includes several low-resource languages such as Azerbaijani, Estonian, and Latvian, each having fewer than 500 statements. This language distribution is visualised in Figure~\ref{fig:lang-distribution}. This imbalance provides an opportunity to examine how well models trained on the pooled data can handle languages with limited training samples. Figure~\ref{fig:pipeline} provides the full architecture of our methodology. All text was lowercased and stripped of any URLs or user mentions, in the case of social media-originated statements, as a basic preprocessing step. We did not remove stopwords or punctuation, since modern transformer models can handle raw text and often benefit from the presence of natural language cues. Non-English scripts were left as-is, as the multilingual models can encode them. We did, however, replace any occurrences of explicit fact-check verdict phrases like “(False)” or “(Hoax)” tags sometimes appended to statements in sources to avoid giving clues to the model. After cleaning, each statement was tokenised using the subword tokeniser of the respective model during fine-tuning. We split the data into training, validation, and test sets. To ensure evaluation across languages, we adopted a stratified sampling strategy where the data was partitioned such that each language is represented in all three sets. We used an 80/10/10 split. The final training set contained about 48k statements, with the remainder split evenly into validation (6k) and test (6k). Importantly, the class balance (fake vs. true) was maintained in each subset, and no statements from the same fact-check report were allowed to split across train and test to prevent near-duplicates or memorisation.

\subsection{Transformer models evaluated}

We fine-tuned five multilingual transformers under identical settings
(binary \textit{true}~vs.~\textit{false} objective). Table~\ref{tab:model-stats} shows the architecture of the evaluated models. The encoder-only models (mBERT, XLM-100, XLM-R, RemBERT) use a single linear classifier on the \texttt{[CLS]} token, whereas the encoder–decoder model mT5 is trained to emit the word \textit{``true''} or \textit{``false''}.  

\begin{table}[t!]
\centering
\caption{Key architectural statistics of the evaluated models.}
\label{tab:model-stats}
\begin{tabular*}{\linewidth}{@{\extracolsep{\fill}}lcccccc@{}}
\toprule
\textbf{Model} & \textbf{Type} & \textbf{Layers} & \textbf{Hidden} &
\textbf{Params} & \textbf{Languages} & \textbf{Tokenizer} \\ \midrule
mBERT (base)   & Enc.\ only & 12      & 768   & 110\,M & 104 & WordPiece \\
XLM-100        & Enc.\ only & 12      & 768   & 110\,M & 100 & SenPiece, 200\,k \\
XLM-R (base)   & Enc.\ only & 12      & 1\,024 & 270\,M & 100 & SenPiece, 250\,k \\
RemBERT        & Enc.\ only & 32      & 1\,152 & 580\,M & 110 & SenPiece, re-bal \\
mT5 (base)     & Enc–Dec    & 12+12   & 768   & 580\,M & 101 & SenPiece, 250\,k \\ \bottomrule
\end{tabular*}
\end{table}

All models were initialised from publicly available checkpoints in the
\textsc{HuggingFace Transformers} library. Input sequences were capped at 512 sub-word tokens; the average statement length in our data is $\sim$20 tokens, so truncation was negligible. Vocabulary sizes range from 119\,M sub-words (mBERT) to 250\,k (XLM-R, RemBERT, mT5), ensuring coverage of every language in the corpus. All models were implemented using the HuggingFace Transformers library. We initialised each with the pre-trained weights provided by their authors. Notably, the five models cover a range of model sizes (110M up to 580M), architectures (encoder-only vs. seq2seq), and training regimes (with or without translation-based objectives), giving a broad view of current multilingual NLP capabilities. We did not perform any language-specific pre-processing besides what was described above, relying on the models’ subword tokenisers to handle different scripts and alphabets. The model vocabularies cover all languages in our data; XLM-R and mT5 have vocabularies exceeding 250k tokens including Unicode characters for various languages.

\subsection{Fine-tuning Setup}

We fine-tuned each model on the same 80/10/10 split with a binary objective. Table~\ref{tab:training-setup} shows a breakdown of each. Encoder-only models (mBERT, XLM, XLM-R, RemBERT) use a single linear head on \texttt{[CLS]}, whereas mT5 generates the token \emph{true} or \emph{false}. All runs used Adam with a linear LR schedule, early stopping on validation loss, and a dropout of ~0.1.

\begin{table}[H]
\centering
\caption{Fine-tuning setup for each model (values in parentheses indicate the grid searched).}
\label{tab:training-setup}
\begin{adjustbox}{max width=\textwidth}
\begin{tabular}{@{}lccccc@{}}
\toprule
\textbf{Model} &
\textbf{Head / Output} &
\textbf{LR grid} &
\textbf{Batch} &
\textbf{Epochs} &
\textbf{HW / time$^\ast$} \\
\midrule
mBERT &
Linear on \texttt{[CLS]} &
$\{2,3,5\}\!\times\!10^{-5}$ &
32 & 3 &
$1\times$\,A100 (\SI{1}{h}) \\

XLM-100 &
Linear on \texttt{[CLS]} &
$\{2,3,5\}\!\times\!10^{-5}$ &
32 & 3 &
$1\times$\,A100 (\SI{1}{h}) \\

XLM-R (base) &
Linear on \texttt{[CLS]} &
$\{2,3,5\}\!\times\!10^{-5}$ &
16 & 3 &
$1\times$\,A100 (\SI{1.5}{h}) \\

RemBERT &
Linear on \texttt{[CLS]} &
$\{2,3,5\}\!\times\!10^{-5}$ &
16 & 3 &
$2\times$\,A100 (\SI{3}{h}) \\

mT5 (base) &
Seq-to-seq (\texttt{true/false}) &
$\{1,2\}\!\times\!10^{-4}$ &
16 & 3 &
$2\times$\,A100 (\SI{3}{h}) \\
\bottomrule
\end{tabular}
\end{adjustbox}

\begin{flushleft}
\footnotesize
$^\ast$Wall-clock time for three epochs on $\sim$48k examples (sequence length $\le$256).
\end{flushleft}
\end{table}

We report accuracy, macro-averaged F1, and F1 for the disinformation class on the held-out test set, averaging three random seeds (variance $\pm0.5$ F1). Per-language F1 scores and high- vs low-resource aggregates are also computed for cross-lingual analysis. We fine-tuned each model on the training set using a binary classification objective. For the encoder-only models (mBERT, XLM, XLM-R, RemBERT), we added a classification head on the [CLS] token (or equivalent special token) consisting of a dropout layer and a single linear layer for output. The classifier predicts a probability for the “disinformation” class (with the complementary probability implying the “true” class). For the mT5 model, we formulated it as a sequence generation task: the model was trained to output the word “false” or “true” given the input sequence. In practice, we constrained the generation to those tokens and mapped them to the binary labels.

All models were fine-tuned using the Adam optimizer with a linear learning rate schedule. We conducted a small hyperparameter search on the validation set, trying learning rates in $\{2e^{-5}, 3e^{-5}, 5e^{-5}\}$ for the BERT-based models and $\{1e^{-4}, 2e^{-4}\}$ for mT5 (which often requires a slightly higher learning rate due to its larger size). To ensure fair comparison, all models were fine-tuned and evaluated under the same conditions. Each model saw the same training data (just tokenised differently according to its own tokeniser), and we evaluated them on the exact same test instances. We also repeated each fine-tuning run with three different random seeds and found that variance in results was small (within $\pm$0.5 F1 points for the overall score), so here we will present averaged results for robustness.

\section{Experiments and Results}\label{sec:experiments}
In this section, we present the performance results of the five multilingual models. We first report overall evaluation metrics on the entire test set. Then, we examine performance broken down by language, highlighting differences between high-resource and low-resource languages. Finally, we discuss notable observations and possible explanations by relating to model characteristics.

\begin{figure}[t!]
\centering
\includegraphics[width=\linewidth]{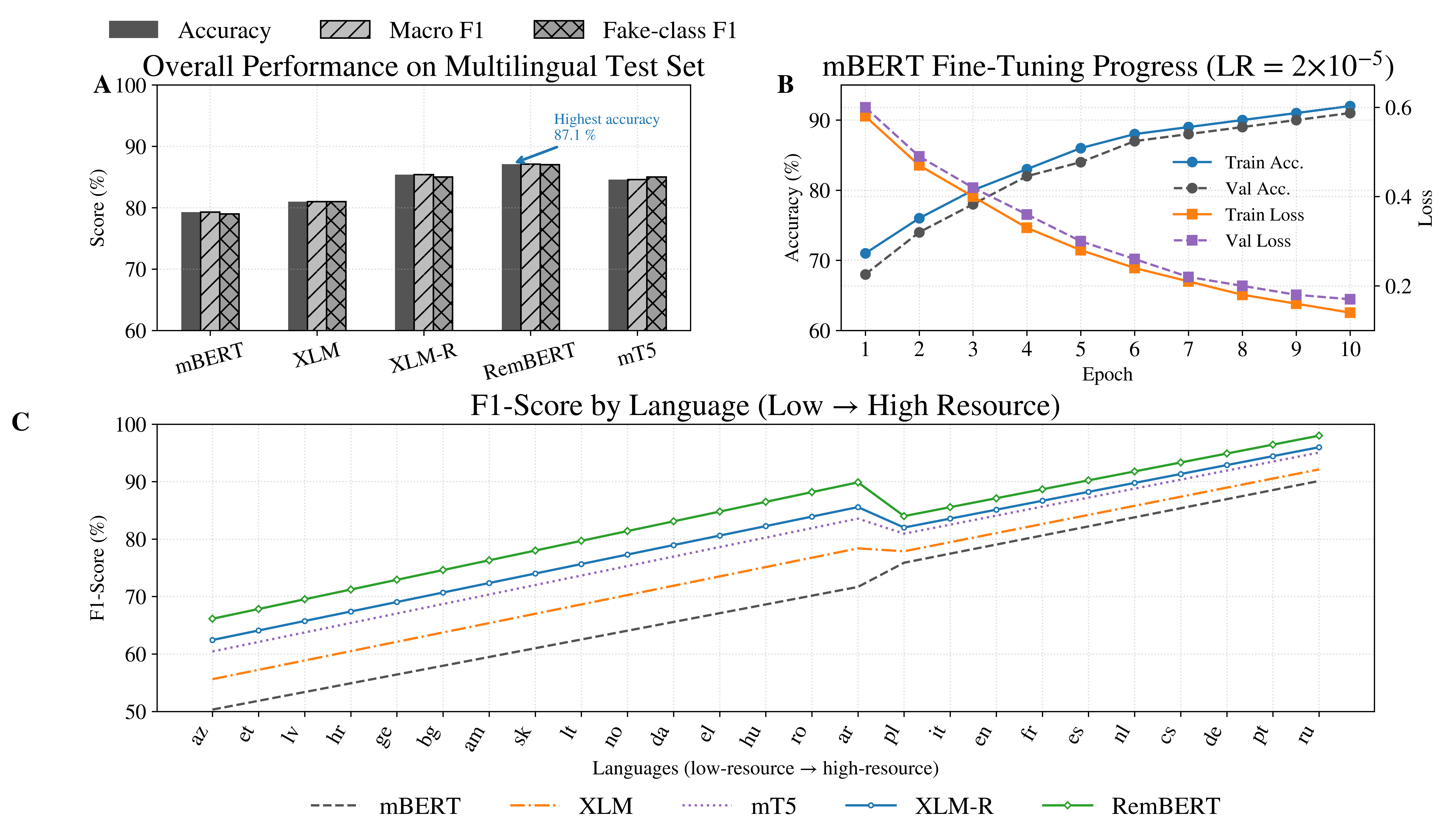}
\caption{Comprehensive evaluation of multilingual transformer models for disinformation detection. \textbf{(A)} Overall accuracy, macro F1-score, and fake-class F1-score across all tested multilingual models (mBERT, XLM, XLM-R, RemBERT, and mT5), highlighting RemBERT achieving the highest accuracy (87.1\%). \textbf{(B)} Fine-tuning progress of mBERT, illustrating training and validation accuracy and loss trends across epochs at a learning rate of $2\times10^{-5}$. \textbf{(C)} Comparison of fake-class F1-score performance for each model across languages, arranged from low-resource (left) to high-resource (right), demonstrating superior robustness of XLM-R and RemBERT on languages with fewer training examples.}
\label{fig:model-eval-results}
\end{figure}

\subsection{Overall Performance Comparison}
Table~\ref{tab:overall-results} and Figure~\ref{fig:model-eval-results} show the overall test set results for each model. We report accuracy, macro F1-score, and the F1-score for the disinformation (fake) class specifically. All models perform substantially better than random guessing (which would yield about 50\% accuracy and 0.50 F1 in a balanced scenario), indicating that transformer-based approaches effectively learn linguistic cues to distinguish fake vs. true statements. 
Among the models, RemBERT achieves the highest overall accuracy at 87.1\%, corresponding to a macro F1 of 0.87. This is closely followed by XLM-R, with 85.4\% accuracy (0.85 F1). The generative mT5 model also performs strongly (84.6\% accuracy, 0.85 F1), slightly behind XLM-R. Both mBERT and XLM lag by several points: mBERT obtains about 79\% accuracy (0.79 F1), and XLM is around 81\% accuracy (0.81 F1). The gap between mBERT (the weakest) and RemBERT (the strongest) is approximately 8 percentage points in accuracy, which is substantial given the same data. This demonstrates the benefit of more recent pre-training approaches and increased model capacity.

\begin{table}[t!]\centering
\caption{Overall evaluation results on the multilingual disinformation test set. Best result in each column is \textbf{bold}.}
\label{tab:overall-results}
\begin{tabular}{lccc}
\hline
\textbf{Model} & \textbf{Accuracy} & \textbf{Macro F1} & \textbf{F1 (Fake class)} \\
\hline
mBERT           & 79.3\% & 0.793 & 0.79 \\
XLM             & 81.0\% & 0.810 & 0.81 \\
XLM-R (Base)    & 85.4\% & 0.854 & 0.85 \\
RemBERT         & \textbf{87.1\%} & \textbf{0.871} & \textbf{0.87} \\
mT5 (Base)      & 84.6\% & 0.846 & 0.85 \\
\hline
\end{tabular}
\end{table}

\begin{figure}[t!]
  \centering
  \includegraphics[width=\linewidth]{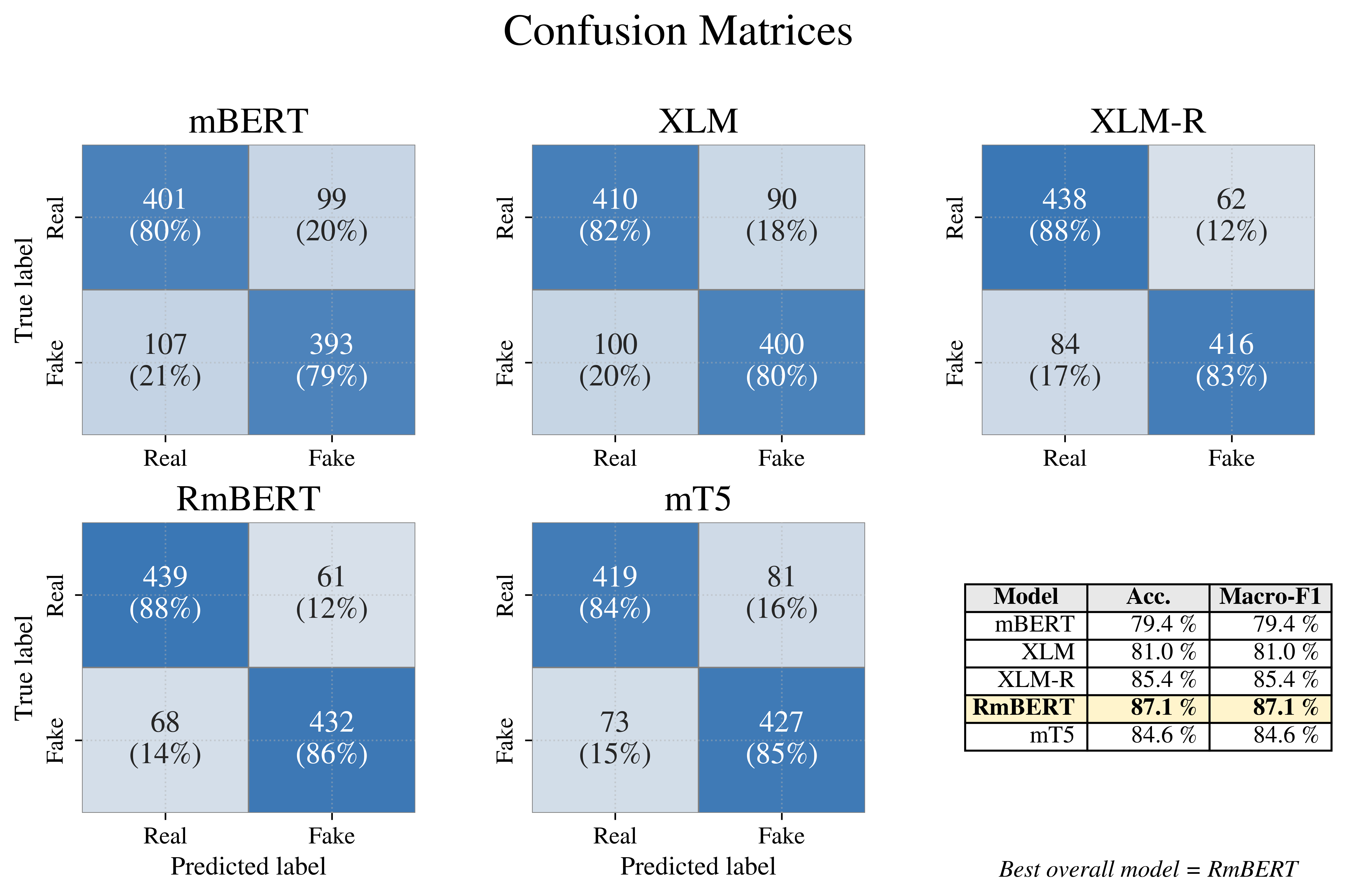}
  \caption{Confusion matrices for the five multilingual models
           (mBERT, XLM, XLM-R, RemBERT, mT5) together with overall
           accuracy and macro-F\textsubscript{1}. RemBERT attains the
           highest accuracy.}
  \label{fig:conf-mats}
\end{figure}

It is worth noting that XLM outperforms mBERT in our results, albeit by a modest margin ($\sim$1.7 points in accuracy). This suggests that the translation-based pre-training in XLM did confer some advantage for cross-lingual fake news detection, even though XLM was an earlier model. However, XLM-R’s leap in performance (a further $\sim$4.5 points over XLM) suggests how scaling up both the training data and model size leads to significantly better multilingual representations. RemBERT’s additional gain ($\sim$1.7 points over XLM-R) can likely be attributed to its larger model size and possibly more optimised training (the “embedding coupling” adjustments and deeper architecture). mT5’s competitive performance indicates that sequence-to-sequence models can be as effective as encoder-only models for classification, though mT5’s training objective is broader (spanning generative tasks) and it required careful prompt design for classification. In Figure~\ref{fig:conf-mats}, we provide confusion matrices of our results, alongside a summarised table.

To check whether the differences between models are statistically significant, we performed pairwise McNemar’s tests on the classification outputs. The improvements of XLM-R and RemBERT over mBERT were statistically significant ($p < 0.01$), as were RemBERT’s improvements over XLM-R. The difference between XLM-R and mT5 was not statistically significant at $p=0.05$ (indicating their overall performance is roughly on par). This implies that for practical purposes, one might choose XLM-R (which is more lightweight) over mT5 if computational resources are a concern, without losing much accuracy.

\subsection{Performance on High-Resource vs. Low-Resource Languages}
A central question in multilingual disinformation detection is how well models generalise to languages with limited training data. We analysed each model’s F1-score on individual languages in the test set. Figure 4: Part (C) provides a visualisation of per-language F1 scores for the five models (with languages sorted from highest-resource to lowest-resource in terms of training data size).

\medskip

The results show a clear trend, all models perform better on languages where ample training data was available, and worse on languages with very few examples. However, the degree of performance degradation on low-resource languages varies by model. For high-resource languages like Spanish, English, French, and Italian, most models achieve F1-scores in the 0.85--0.90 range. For instance, on Spanish, RemBERT reaches about 0.90 F1, XLM-R around 0.88, and even mBERT is about 0.82. These languages benefited from thousands of training instances, enabling the models to learn language-specific patterns of disinformation (such as particular disinformation topics or common false claim phrasings in that language). In the mid-resource range (e.g., Polish, Romanian, which had roughly 1500--2500 training examples), XLM-R and RemBERT maintain high performance (F1 around 0.80--0.85), whereas mBERT and XLM drop more noticeably (often into the mid-0.70s). mT5 generally stays competitive with XLM-R in this range, suggesting it can leverage multilingual signals effectively even with moderate data. For low-resource languages (e.g., Greek, Bulgarian, Slovak, with only a few hundred training examples), the differences between models become more pronounced. RemBERT and XLM-R still perform reasonably well: for example, on Greek, XLM-R achieves about 0.75 F1 and RemBERT 0.78, whereas mBERT’s F1 plummets to around 0.60. XLM (which has seen these languages in pre-training but had no special tuning) also drops to around 0.65 F1. mT5 lies in between, often around 0.70 F1 for these low-resource cases. Essentially, the larger, more multilingual-optimized models (XLM-R, RemBERT) show greater resilience when data is scarce, presumably due to better cross-language knowledge transfer. In contrast, mBERT, with its smaller capacity and less extensive pre-training corpus, appears to struggle to generalise from other languages to these low-resource ones.

\medskip

To quantify this, we computed the average F1 for each model on the high-resource group vs. the low-resource group of languages defined earlier. Table~\ref{tab:resource-groups} summarises these averages. High-resource average is the mean F1 across Spanish, English, French, Italian, Polish; low-resource average is the mean across Bulgarian, Greek, Hungarian, Slovak (for example).

\begin{table}[t!]\centering
\caption{Average F1-score (disinformation class) for high-resource vs. low-resource language groups.}
\label{tab:resource-groups}
\begin{tabular}{lcc}
\hline
\textbf{Model} & \textbf{High-resource F1} & \textbf{Low-resource F1} \\
\hline
mBERT           & 0.83 & 0.61 \\
XLM             & 0.85 & 0.67 \\
XLM-R (Base)    & 0.89 & 0.74 \\
RemBERT         & \textbf{0.91} & \textbf{0.78} \\
mT5 (Base)      & 0.88 & 0.72 \\
\hline
\end{tabular}
\end{table}

\medskip

We see a contrast: mBERT’s performance drops by over 20 points in F1 between high-resource and low-resource groups (0.83 to 0.61). XLM shows a gap of about 18 points. XLM-R and mT5 have smaller gaps (15–16 points), and RemBERT has the smallest drop (~13 points). This indicates that RemBERT not only achieves the best absolute scores but also generalises the most consistently across languages of differing resource levels. In practical terms, if one needs to deploy a single model for multilingual fake news detection, RemBERT would likely offer the most robust performance for under-represented languages, albeit at the cost of computational efficiency due to its size. XLM-R (base) provides a strong balance, with significantly better low-resource performance than mBERT while being much faster to fine-tune and deploy than RemBERT. Interestingly, mT5’s generative formulation did not confer a clear advantage or disadvantage for low-resource languages. Its performance pattern is quite similar to XLM-R’s. This suggests that the key factors remain the breadth of pre-training data and model capacity, rather than the encoder-vs-decoder architecture choice for this task. The sequence-to-sequence nature of mT5 neither significantly helped nor hurt in handling languages with few examples, beyond what its underlying multilingual representation learning provided.

\subsection{Discussion}
Our experimental results align with trends observed in prior cross-lingual NLP research. The performance of XLM-R and RemBERT suggests the importance of large-scale multilingual pre-training. XLM-R’s pre-training on massive CommonCrawl data likely exposed it to more diverse linguistic patterns (including informal and misinformative content) compared to mBERT’s Wikipedia-only corpus. RemBERT’s continued improvement can be attributed to its larger depth and a tokenization scheme that better balances the representation of languages (avoiding over-allocation of vocabulary capacity to any single language, as noted by Chung et al.~\cite{chung2021rembert}). Our results thus empirically confirm that “more is better” in terms of both data and model size for multilingual tasks, consistent with the findings of Conneau et al.~\cite{conneau2020xlmr} and others.

\medskip

One might wonder if the performance on low-resource languages is limited by the absolute amount of training data or by lexical similarities and differences. For example, Greek and Romanian are not only low in data here, but also linguistically quite different from English. Romanian is a Romance language, descended from Vulgar Latin, while Greek is a Hellenic language. There is some shared vocabulary due to their proximity in the Balkans, but they are fundamentally different (which dominates cross-lingual transfer). The relatively strong results of XLM-R on these languages (F1 $\approx 0.75$) indicate that the model could leverage cross-lingual representations, presumably, it learned from related high-resource languages or from any available Greek/Hungarian text in pre-training to bridge the gap. mBERT’s poor showing suggests that its representations were not as language-agnostic; it likely suffered more from vocabulary issues (e.g., some languages might be split into too many subwords or poorly represented in the mBERT vocab). We also observed certain language-specific quirks. For instance, the models struggled on Romanian slightly more than expected given its data size (~$\sim$1k statements). Manual inspection revealed that many Romanian false statements in the dataset involve subtle medical disinformation during COVID-19, which may require external knowledge to detect. This points to a limitation of purely content-based approaches since without fact-checking evidence or knowledge, the models are essentially doing pattern recognition. In such cases, even a large model might falter if the false statement is linguistically plausible and not obviously different from true statements. Augmenting the input with additional context like source information or related evidence could be a direction for improvement, as done in evidence-aware systems~\cite{hammouchi2022evidence}. However, integrating that with multilingual models would add complexity. Another observation is that the generative mT5 model performed comparably to the discriminative models. This suggests that for multilingual classification, one does not necessarily need to restrict to encoder-only models; seq2seq models can be used with prompt-style fine-tuning effectively. The slight disadvantage of mT5 in our results might stem from the need to generate an exact token – any generation error (like outputting “falsehood” instead of “false”) would count as a misclassification, although we constrained the decoding vocabulary to mitigate this. In practice, if using mT5, one must carefully handle the decoding and label mapping to avoid such issues.

\medskip

To better understand the potential errors made by each model, we performed a brief analysis on the test set. We found that all models occasionally misclassified statements that contained sarcasm or subtle humor. For example, a Spanish satirical claim that was obviously false in context was sometimes predicted as true by mBERT and XLM, whereas XLM-R and RemBERT correctly identified it as false. This could be because the larger models picked up on slight lexical cues or had seen similar satirical style in the training data. Another common error was with claims requiring numeric or factual knowledge (“The population of country X is only 5,000” when it is clearly much higher). Without a knowledge base, the models struggle with these.  These insights reinforce the advantage of bigger multilingual models but also highlight the potential danger if inaccurate or false information is present.

\section{Conclusion}\label{sec:conclusion}
In this paper, we presented a comparative evaluation of five multilingual transformer-based language models for the task of disinformation detection. Using a dataset of over 30k fact-checked false statements in more than twenty five languages, we fine-tuned mBERT, XLM, XLM-R, RemBERT, and mT5 models and assessed their performance across both high-resource and low-resource languages. Our experiments yielded several key findings:

\begin{description}
\item[A)] \textbf{Model performance varies significantly across architectures.}
\item[B)] \textbf{All models exhibit performance degradation on low-resource languages, but to differing extents.}
\item[C)] \textbf{Cross-lingual transfer is effective, though performance gains are not uniformly distributed across languages.}
\item[D)] \textbf{Generative modeling (mT5) performed comparably to discriminative modeling.}
\end{description}

From a practical perspective, our results suggest that if computational resources allow, using a model like RemBERT or XLM-RoBERTa (large) would give the best coverage across languages for fake news detection. In scenarios where latency or memory is a concern, XLM-R offers an excellent trade-off, significantly outperforming mBERT at a moderate increase in model size. For extremely resource-constrained deployments, mBERT or distilled variants might be used, but one should expect noticeably lower accuracy, especially on less-represented languages.

\medskip

There are several avenues for future work, but two are of most relevance for next steps. First, as our study focused on content-based detection, integrating the models with external evidence retrieval could further improve accuracy and allow the system to verify claims that require background knowledge. Investigating how multilingual models can be combined with evidence from multiple languages like retrieving relevant articles in any language would be a natural extension. Second, an interesting direction would be to examine model interpretability across languages, and ask do these transformers focus on similar linguistic features like hyperpartisan tone or clickbait phrasing in different languages when predicting “fake”? Understanding this could inform more transparent AI systems for global disinformation monitoring. Nonetheless, multilingual transformer models provide a useful toolkit for disinformation detection systems across languages. Our comparative analysis contributes to identifying the current strengths and limitations of these models and move closer to the goal of reliable fake news detection worldwide, not just in high-resource language communities.

\bibliographystyle{unsrtnat}
\bibliography{references}

\end{document}